\documentclass[runningheads]{llncs}
\usepackage[T1]{fontenc}
\usepackage{graphicx}
\usepackage{siunitx}
\usepackage{multirow}
\usepackage{tabularx,booktabs}
\usepackage{array}
\usepackage[table,xcdraw]{xcolor}
\usepackage{caption}
\usepackage{geometry}
\usepackage{arydshln}
\usepackage{subcaption}
\begin{document}
\title{From Neurons to Computation: \\ Biological Reservoir Computing for Pattern Recognition}
%
%
\author{Ludovico Iannello\inst{1}\textsuperscript{*}\orcidID{0009-0003-1392-7307} \and
Luca Ciampi\inst{1}\textsuperscript{*}\orcidID{0000-0002-6985-0439} \and
Gabriele Lagani\inst{1}\textsuperscript{*}\orcidID{0000-0003-4739-5778} \and
Fabrizio Tonelli\inst{2}\orcidID{0000-0003-4167-555X} \and
Eleonora Crocco\inst{2} \and
Lucio Maria Calcagnile\inst{3}\orcidID{0000-0003-3572-6154} \and
Angelo Di Garbo\inst{3} \and
Federico Cremisi\inst{2}\orcidID{0000-0003-4925-2703} \and
Giuseppe Amato\inst{1}\orcidID{0000-0003-0171-4315}
}
\authorrunning{L. Iannello, L. Ciampi, G. Lagani et al.}
%
\institute{Institute of information science and technologies (CNR-ISTI), Pisa, Italy \and
Biology Laboratory of Scuola Normale Superiore (Bio@SNS), Pisa, Italy \\
\and
Institute of biophysics (CNR-IBF), Pisa, Italy
}
\maketitle              

\renewcommand{\thefootnote}{\fnsymbol{footnote}}
\footnotetext[1]{Corresponding authors, they contributed equally to this work: \texttt{ludovico.iannello@isti.cnr.it}, \texttt{luca.ciampi@isti.cnr.it}, , \texttt{gabriele.lagani@isti.cnr.it}}
\renewcommand{\thefootnote}{\arabic{footnote}}

\begin{abstract}
In this paper, we introduce a paradigm for reservoir computing (RC) that leverages a pool of cultured biological neurons as the reservoir substrate, creating a \textit{biological reservoir computing} (BRC). This system operates similarly to an echo state network (ESN), with the key distinction that the neural activity is generated by a network of cultured neurons, rather than being modeled by traditional artificial computational units. The neuronal activity is recorded using a multi-electrode array (MEA), which enables high-throughput recording of neural signals. In our approach, inputs are introduced into the network through a subset of the MEA electrodes, while the remaining electrodes capture the resulting neural activity. This generates a nonlinear mapping of the input data to a high-dimensional biological feature space, where distinguishing between data becomes more efficient and straightforward, allowing a simple linear classifier to perform pattern recognition tasks effectively.
To evaluate the performance of our proposed system, we present an experimental study that includes various input patterns, such as positional codes, bars with different orientations, and a digit recognition task. The results demonstrate the feasibility of using biological neural networks to perform tasks traditionally handled by artificial neural networks, paving the way for further exploration of biologically-inspired computing systems, with potential applications in neuromorphic engineering and bio-hybrid computing.

\keywords{Reservoir Computing  \and Bio-Inspired Neural Networks \and Multi-Electrode Array \and Neural Computing.}
\end{abstract}

\section{Introduction}
\label{sec:intro}
Reservoir Computing (RC)~\cite{DBLP:journals/csr/LukoseviciusJ09} is a powerful machine learning paradigm that leverages a highly nonlinear dynamical system with numerous degrees of freedom to project input data into a high-dimensional latent space. In this space, data samples are often more easily linearly separable, which facilitates efficient downstream tasks such as classification or regression. Several models have been proposed to implement the reservoir, among which the Liquid State Machine (LSM)~\cite{DBLP:journals/neco/MaassNM02} and the Echo State Network (ESN)~\cite{jaeger2001,DBLP:journals/corr/abs-1712-04323} are the most prominent. The LSM employs a pool of nonlinear spiking neural units to generate a dynamic representation of input signals, while the ESN uses continuous-valued neurons and does not aim to replicate biologically realistic spiking behavior. These models have enabled successful applications in various domains, including time-series prediction~\cite{DBLP:journals/tnn/BianchiSLJ21} and speech recognition~\cite{DBLP:journals/ficn/YonemuraK24}.


\begin{figure*}[t]
    \centerline{\includegraphics[width=0.9\linewidth]{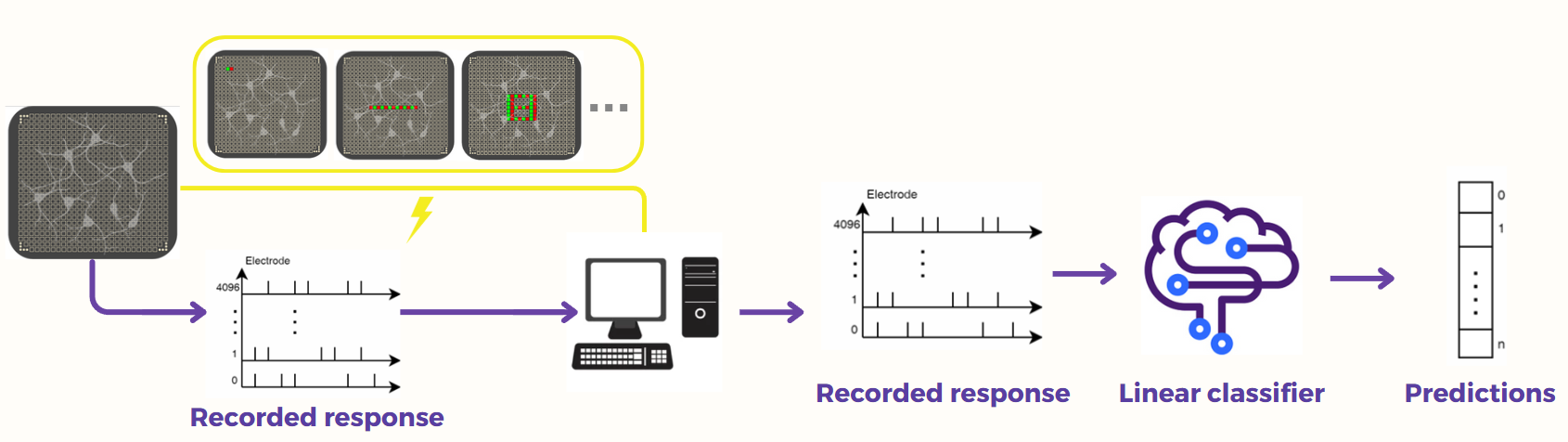}}
    \caption{\textbf{Overview of our biological reservoir computing (BRC) approach.} A multi-electrode array (MEA) serves as an interface to both stimulate and record activity from a cultured biological neural network. Each input is mapped to a specific subset of MEA electrodes, which deliver the corresponding stimulus to the network. The resulting spiking activity is recorded by a different set of electrodes and assembled into a vector that represents the input in a high-dimensional latent space. The complex dynamics of the BRC system ensure that this mapping is highly nonlinear. Finally, a linear classifier is trained to recognize the category of the original input based on its latent representation.}
    \label{fig:teaser}
\end{figure*}

In this paper, we introduce an extension to the RC paradigm: the use of a cultured network of real biological neurons as the reservoir substrate. This approach, which we refer to as the \textit{biological reservoir computing} (BRC), employs a \textit{biological} dynamical system to map input samples into a high-dimensional space of biological features. Unlike traditional RC systems, where the reservoir is typically composed of artificial neurons or units, our biological system leverages real neurons grown \textit{in vitro}, providing a natural bridge between computational and biological systems. This paradigm has the potential to improve energy efficiency -- a significant limitation of current deep learning technologies~\cite{DBLP:conf/aiia/BadarVSGMIAZ21,JAVED2010907} -- while also advancing our understanding of biological neuron behavior. Indeed, neuroscientific observations suggest that transient dynamics in the brain carry a significant portion of the input information, indicating that some form of RC may play a role in high-level biological cognition~\cite{https://doi.org/10.1111/j.1460-9568.2007.05976.x}.
In more detail, we derive our neurons from stem cells using well-established neurobiological techniques~\cite{Chambers_2009,PMID:33139941}. These cultured networks are then interfaced with a high-density multi-electrode array (MEA) device~\cite{bonifazi}, which contains 4096 electrodes for both electrical stimulation and neural response recording (see also Fig.~\ref{fig:teaser} for a visual overview of our approach). 

To evaluate the ability of our BRC system to generate discriminative feature representations across varying input complexities, we conduct an experimental study spanning tasks of several difficulties. Specifically, we feed features derived from biological culture responses into a single-layer perceptron and assess classification performance of the combined system (i.e., our BRC system followed by the perceptron). We begin with simple pointwise stimuli as basic input patterns, then progress to more complex geometric stimuli such as bars with varying orientations. Finally, we evaluate digit recognition to test the utility of the BRC system in real-world classification scenarios. Results are promising, indicating that biologically cultured networks can function as effective substrates for reservoir computing, producing high-dimensional features that improve pattern recognition accuracy.

In summary, the contributions of this paper are as follows: 

\begin{itemize} 
    \item We introduce a reservoir computing paradigm that leverages biologically cultured networks as substrates for generating high-dimensional feature representations, which we term biological reservoir computing (BRC). Our approach interfaces with these cultured networks using a high-density multi-electrode array (MEA) device, enabling both electrical stimulation and the recording of neural responses. 
    \item We define and calibrate a stimulation protocol for interacting with neurons via the MEA device, ensuring robust and consistent neuronal responses across different stimulation parameters. 
    \item We present an experimental investigation to validate the proposed BRC system in several pattern recognition tasks, assessing the efficacy of the biological feature representations in distinguishing various categories of input patterns. 
\end{itemize}

The remainder of this manuscript is structured as follows: Sec.~\ref{sec:rel_work} provides an overview of related work in the field of reservoir computing and biological computing; Sec.~\ref{sec:method} details the proposed methodology, including the neurobiological techniques and MEA interfacing; Sec.~\ref{sec:exp} discusses the experimental setup, results, and analysis; and finally, Sec.~\ref{sec:conclusions} presents our conclusions and outlines future directions.

\section{Related Work} 
\label{sec:rel_work}
A central theme in deep learning (DL) research is the biological plausibility of current methods, with several studies questioning whether existing architectures reflect the complexity and functionality of biological neural systems~\cite{DBLP:journals/ficn/MarblestoneWK16,Hassabis_2017,f3a024ff8e474a1ea4e798aaa5536830,DBLP:conf/atal/Tenenbaum18}. In response, there has been a growing interest in exploring biologically inspired alternatives to push the boundaries of machine learning and cognitive computation~\cite{DBLP:journals/corr/abs-2307-16236,DBLP:journals/corr/abs-2307-16235}. A wide range of models has been proposed to bridge this gap, either by adopting biologically realistic neural computations~\cite{DBLP:journals/ficn/Diehl015,DBLP:journals/ficn/FerreMT18,DBLP:conf/ner/LaganiMFGCPCA21,DBLP:journals/cogcom/SunCCS23} or by refining synaptic dynamics and learning mechanisms~\cite{DBLP:journals/pnas/KrotovH19,DBLP:journals/nn/IllingGB19,DBLP:conf/mod/LaganiFGA21,DBLP:journals/nca/LaganiFGA22,DBLP:conf/cbmi/LaganiBGFGA22,DBLP:conf/sisap/LaganiGFA22,DBLP:conf/iclr/JourneRGM23,DBLP:journals/ijon/LaganiFGFA24,hebbian_eccv_workshop,DBLP:journals/corr/abs-2412-03192}.

Reservoir Computing (RC)~\cite{DBLP:journals/csr/LukoseviciusJ09,DBLP:journals/cogcom/ScardapaneBBM17} represents one of the bio-inspired paradigms that has gained considerable attention for its potential to model complex neural dynamics. Specifically, Echo State Networks (ESNs)~\cite{DBLP:journals/corr/abs-1712-04323,jaeger2001} and Liquid State Machines (LSMs)~\cite{DBLP:journals/neco/MaassNM02,DBLP:journals/tnn/ZhangLJC15} are two prominent models within the RC framework. While LSMs leverage biologically inspired spiking neuron models~\cite{DBLP:books/cu/GerstnerK02,PhysRevE.48.1483} to implement highly nonlinear neural dynamics, ESNs rely on continuous-valued neurons and random recurrent connectivity to generate rich internal dynamics. In both cases, the reservoir transforms input signals into a high-dimensional feature space, facilitating pattern separation for downstream tasks. Although the connectivity is typically random, specific algorithms are employed to ensure that the network dynamics remain stable and computationally effective~\cite{DBLP:journals/corr/abs-1712-04323,DBLP:conf/inista/SarliGM20}.
Several extensions and related models have been proposed to enhance the biological plausibility of reservoir computing systems. For instance, some variants of ESNs~\cite{DBLP:conf/icpr/WangL16} incorporate biologically inspired mechanisms such as spike-timing-dependent plasticity (STDP)~\cite{Song_2000,DBLP:books/cu/GerstnerK02}, which adjusts synaptic strengths based on the timing of pre- and post-synaptic activity. Another biologically motivated model, the Self-Organizing Recurrent Network (SORN)~\cite{DBLP:journals/ficn/LazarPT09}, combines STDP with homeostatic plasticity~\cite{DBLP:conf/ijcnn/CarlsonRDK13} to maintain network stability and support unsupervised learning in recurrent architectures. Additionally, a gating mechanism proposed in~\cite{DBLP:conf/inista/SarliGM20} enhances ESN performance by improving the recall of long-term dependencies, which are often degraded in highly nonlinear recurrent systems.
A related technique, the Extreme Learning Machine (ELM)~\cite{DBLP:journals/ijon/HuangZS06,DBLP:journals/cogcom/Huang14}, employs a large pool of feedforward nonlinear neurons with randomly initialized weights to project inputs into a high-dimensional feature space. Although ELMs lack recurrent connections and differ architecturally from RC models, they share the core principle of using random nonlinear mappings to facilitate efficient learning.
These biologically inspired and computationally efficient models have demonstrated success in various applications, including speech processing~\cite{DBLP:journals/tnn/ZhangLJC15} and continual learning~\cite{DBLP:conf/esann/CossuBCGL21}, among others.

In contrast to previous works, this contribution enhances the biological plausibility of RC approaches by leveraging a real biological neural network as the computational reservoir. Specifically, we introduce the concept of biological reservoir computing (BRC), in which cultured biological neurons serve as the reservoir substrate. Prior studies have explored interfacing with biological neurons via microelectrode array (MEA) devices~\cite{Shahaf8782,DBLP:journals/tbe/RuaroBT05,DBLP:journals/ijon/FerrandezLPF13,DBLP:journals/ploscb/IsomuraKJ15,GOEL2016320,DBLP:journals/ploscb/PastoreMGM18,KAGAN20223952}, but our work specifically investigates the use of cultured neuronal networks as a potential RC substrate. A few previous efforts have also employed biological neurons in RC frameworks; however, they either focused on input separation through distinct stimuli~\cite{DOCKENDORF200990} or relied on alternative interfacing methods such as optogenetics and calcium imaging instead of MEA-based systems~\cite{Sumi202306}. Our approach offers a promising direction for advancing the field by integrating biological computation with machine learning methodologies, potentially yielding benefits such as improved energy efficiency and deeper insights into the dynamics of biological neural systems.

\section{Reservoir of Biological Neurons} 
\label{sec:method}

\begin{figure}[t]
    \centerline{\includegraphics[width=0.5\linewidth]{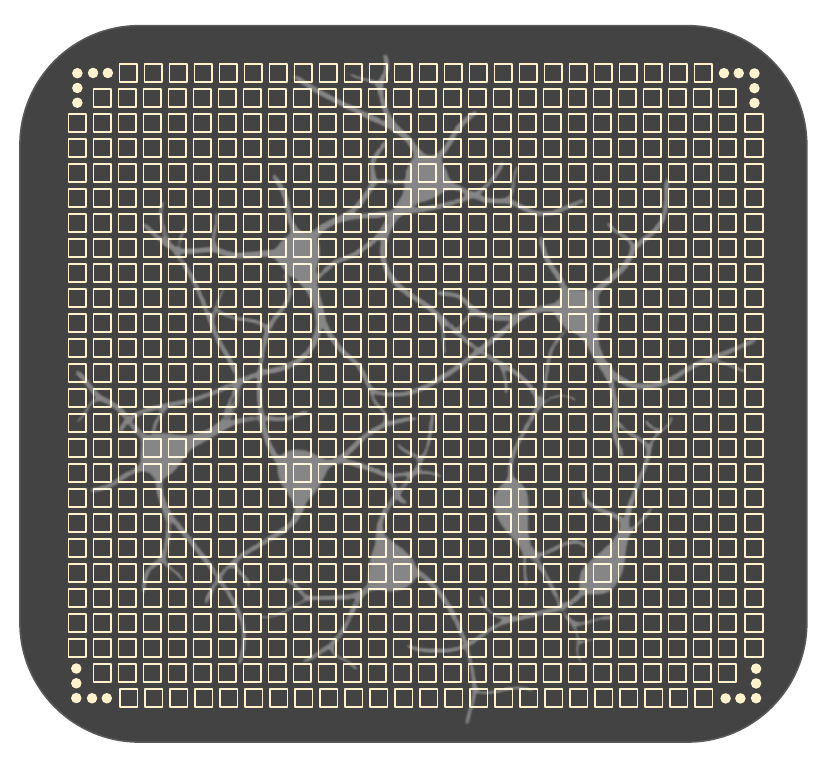}}
    \caption{\textbf{Schematic illustration of a cultured biological network interfaced with the MEA device.} Each square represents an individual MEA electrode. Input patterns are mapped onto the MEA by assigning elements of the patterns to specific electrodes. Electrical pulses are delivered based on the corresponding input intensities, and the evoked activity of the network is recorded via the remaining electrodes. The resulting spiking responses are used to construct a high-dimensional representation of the input in the feature space of the biological reservoir. Neuron scale in the figure is not to scale and is adjusted for visibility.}
    \label{fig:mea}
\end{figure}

A network of biological neurons can be conceptualized as a collection of randomly interconnected computing units that exhibit complex, nonlinear dynamics. When such a network is cultured on a multi-electrode array (MEA) device -- a grid of electrodes capable of both electrical stimulation and activity recording (see Fig.~\ref{fig:mea}) -- it becomes possible to deliver controlled input signals and monitor the corresponding network responses.

To utilize the biological network as a reservoir, we establish a direct mapping between each input sample and a predefined subset of MEA electrodes. For example, when processing an image, its $H \times W$ grid of pixels is mapped onto an $H \times W$ subset of the electrode array. Each pixel is associated with a specific electrode, and its intensity determines the stimulation parameters used to elicit a neural response. Once the electrical stimulus is delivered, we record the evoked spiking activity from the remaining electrodes on the array. The resulting spike trains are aggregated into a single high-dimensional vector that represents the high-dimensional latent space of the input sample. Thus, in this configuration, the biological reservoir functions as a biologically grounded feature extractor, or \textit{bioware}, that transforms structured input signals into nonlinear, high-dimensional representations~\cite{PMID:33139941,DBLP:journals/tbe/RuaroBT05}. 

To assess the quality of these representations, we feed the extracted features into a single-layer perceptron and evaluate the classification performance of the entire pipeline. A schematic overview of the system is provided in Fig.~\ref{fig:teaser}. The following sections provide details on the stimulation protocol, as well as the training and testing procedures concerning the single-layer perceptron classifier used in our experimental study.

\subsection{Stimulation Protocol}
Interfacing with the biological network presents unique challenges, particularly the need to calibrate multiple stimulation parameters -- such as pulse amplitude, frequency, and duration -- which must be carefully optimized to ensure robust and consistent neuronal responses. To address these challenges, we investigate and define optimal stimulation protocols aimed at facilitating efficient neuronal activity and enabling reliable data recording.

Specifically, to interface with the cultured neural network, specific MEA electrodes are selected for delivering electrical stimulation. For bipolar stimulation, designated electrodes are assigned as positive and negative poles. Each electrical pulse delivered to the network is configured as a rectangular waveform -- either monophasic or biphasic -- with a specified amplitude $\si{\ampere}$ (measured in $\si{\micro\ampere}$) and pulse width $\delta_+$ or $\delta_-$ (measured in $\si{\micro\second}$) for the positive and negative phases, respectively.
Because the MEA distributes the total stimulation current across all selected electrode pairs, the amplitude $\si{\ampere}$ is interpreted on a per-pair basis. 
Stimulation is applied as a sequence of $N$ identical pulses, spaced by a fixed inter-pulse interval $T$ (in seconds).
When configuring the stimulation parameters, we carefully balanced signal strength and hardware longevity. The stimulus must be strong enough to reliably elicit measurable neuronal responses, yet not so intense or repetitive as to risk damaging the electrodes -- an issue observed during preliminary trials involving high-intensity or high-frequency stimulation.

Throughout the stimulation protocol, network activity was recorded continuously. To evaluate the efficacy of the stimulation parameters, we developed a visualization tool that allows for the assessment of stimulation quality. The tool represents the $64 \times 64$ MEA electrode grid as an image, where each pixel corresponds to a specific electrode and the recorded signals are processed using the precise spike time detection (PTSD) algorithm~\cite{MACCIONE2009241}, a widely-used method for accurately extracting spiking activity from extracellular recordings. Specifically, for the stimulus given at time $t_s$, for each electrode $(i, j)$, the tool calculates the activity $a$ during a temporal chunk $C$ of the recorded activity, defined as the number of spikes observed during this chunk: 

\begin{equation} 
    a^C_{ij}(t_s) = \sum_{t=t_s}^{t_s+C} s_{ij}(t).
\end{equation}

Here, $s_{ij}(t)$ is a binary variable indicating whether a spike was recorded at time $t$, and the sum runs over discrete time steps, determined by the recording device’s frame rate, which in our case was the maximum available rate of $f_s = 20,000$ Hz. To quantify the response of the network to each stimulus, we define the response $r$ on a given electrode as the difference between the activity in a temporal chunk $C$ immediately following the stimulus and the activity in a chunk $-C$ immediately preceding it:

\begin{equation} 
   r^C_{ij}(t_s) = a^C_{ij}(t_s) - a^{-C}_{ij}(t_s).
\end{equation}

\begin{figure}[t]
    \centerline{\includegraphics[width=0.5\linewidth]{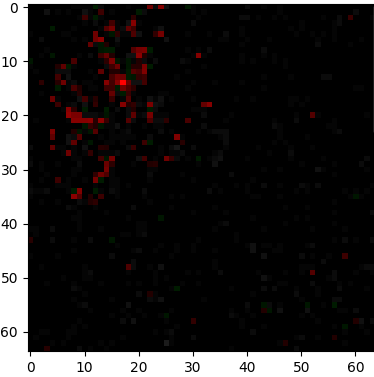}}
    \caption{\textbf{Visualization of the average network response to $N$ stimuli over a temporal chunk $C$.} Each pixel represents an electrode in the $64 \times 64$ MEA grid, with the color intensity encoding the average level of response following the stimuli. We set $C = \SI{10}{\milli\second}$ and $N = 25$.}
    \label{fig:mea_example}
\end{figure}

The tool then visualizes the response of each electrode to a given stimulus by encoding it into the color intensity of the corresponding pixel. Specifically, since we have $N$ recordings (one for each stimulus), we take the average of $r^C_{ij}$ over the $N$ repetitions. In addition to the average, we compute the standard deviation of the observations and estimate confidence intervals for the response using a Student’s t-estimator.
Red pixels indicate a positive response (increased activity) at a confidence level above 99\%, green indicates a positive response at a 95\% confidence level, cyan denotes a significant negative response (decreased activity) at a 95\% confidence level, and yellow marks a negative response at a 99\% confidence level. 
Figure~\ref{fig:mea_example} shows a representative example of the network response visualized using our tool.

\subsection{Classifier Training Procedure}
The training phase of the single-layer perceptron classifier begins by selecting a collection of input patterns to be used for network stimulation through the MEA interface. Each pattern is mapped to a predefined subset of MEA electrodes designated for stimulation. For every input, the corresponding electrodes are activated via electrical pulses repeated $N = 25$ times 
to obtain a distribution of responses suitable for statistical analysis. To mitigate potential carry-over effects and ensure reliable neuronal responses, a $T =  \SI{10}{\second}$ interval is enforced between successive stimuli, allowing the cultured network to return to its resting state~\cite{DBLP:journals/tbe/RuaroBT05}.
We set a temporal chunk $C = \SI{10}{\milli\second}$ -- corresponding to approximately two synaptic transmission cycles~\cite{annurev:/content/journals/10.1146/annurev.physiol.61.1.521} -- and the number of spikes detected on each electrode within this window is counted. This results in a 4096-dimensional feature vector representing the input pattern in the latent space defined by the biological reservoir. To test its effectiveness in transforming the input and propagating information from the stimulated neurons to the others, we excluded a square region around the stimulated area.

Once latent representations for all input patterns have been collected, a linear classifier is trained to associate each feature vector with its corresponding input category. For each class, the recordings are randomly shuffled and split into two independent sample sets for training and testing -- we refer to $N_{training}$ and $N_{testing}$ to express their cardinalities, where $N = N_{training} + N_{testing}$. Specifically, we set $N_{training} = 20$ and $N_{testing} = 5$.
Classification is performed using a single-layer perceptron trained via stochastic gradient descent (SGD)~\cite{10.5555/304710.304720}, minimizing the cross-entropy loss function~\cite{DBLP:journals/nca/KlineB05}. No separate validation set is used, as no early stopping is performed. This decision maximizes the amount of data available for both training and evaluation. The classifier is trained for 20 epochs, a value determined empirically to ensure convergence of the learning process. 
Finally, the trained model is used in the evaluation phase to assess the performance of the biological reservoir as a high-dimensional feature extractor.

\subsection{Classifier Testing Procedure}
To evaluate the performance of the trained model, we use the remaining $N_{testing} = 5$ network activity responses for each input pattern that were withheld during training. The linear classifier is kept fixed in the state achieved at the final training epoch, without further updates or fine-tuning.

Each test sample is acquired following the same protocol used during training: the input pattern is mapped onto the MEA electrodes, electrical stimulation is delivered accordingly, and the resulting network activity response is recorded. This activity is then processed using the same spike detection algorithm~\cite{MACCIONE2009241} to extract precise spiking events. From the extracted spike trains, we compute the number of spikes recorded on each electrode within a temporal window of duration $C = \SI{10}{\milli\second}$, immediately following the stimulus. This spike count vector serves as the latent representation of the test sample.

Because the testing procedure mirrors the training methodology, the distribution of test features remains consistent with that of the training data, supporting a fair and meaningful evaluation. Each test representation is passed through the trained classifier, and the resulting predictions are compared against the ground truth. We compute classification accuracy -- defined as the ratio of correct predictions to the total number of samples -- as the primary metric for assessing the effectiveness of our biologically-grounded reservoir computing system.

\section{Experimental Evaluation} 
\label{sec:exp}

\begin{figure}[t]
    \centering
    \begin{subfigure}{0.8\linewidth}
        \centering
        \includegraphics[page=1, width=\linewidth]{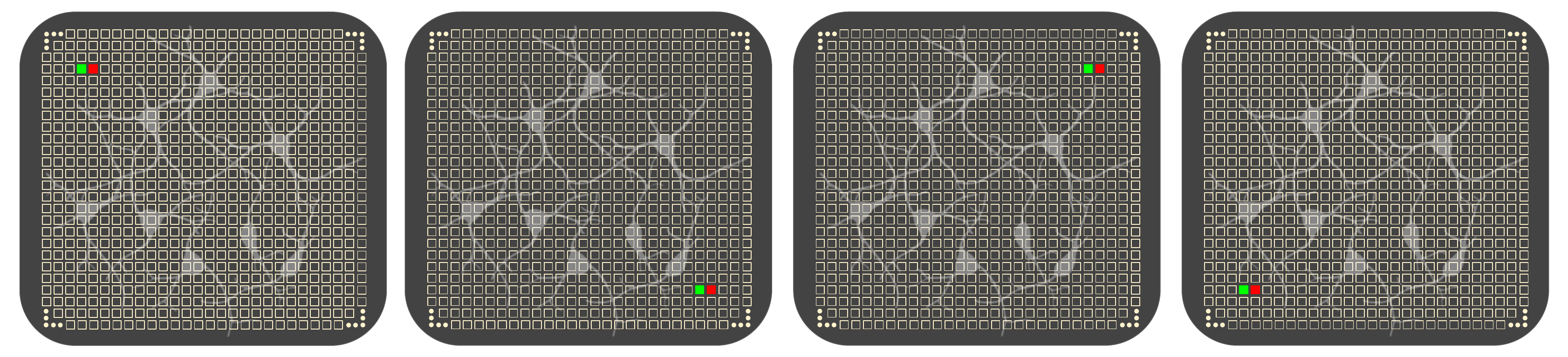}
        \caption{Pointwise stimuli.}
    \end{subfigure}
    \vspace{0.5em}

    \begin{subfigure}{0.8\linewidth}
        \centering
        \includegraphics[page=2, width=\linewidth]{images/images_experiments.pdf}
        \caption{Oriented bars.}
    \end{subfigure}
    \vspace{0.5em}

    \begin{subfigure}{0.8\linewidth}
        \centering
        \includegraphics[page=3, width=\linewidth]{images/images_experiments.pdf}
        \caption{Digit recognition.}
    \end{subfigure}
    \caption{\textbf{Visual representation of the input patterns used across the three experimental scenarios.} The first row displays the pointwise stimuli experiments, where single pairs of adjacent electrodes were stimulated. The second row shows the oriented bar experiments with four categories of bars at 0, 45,  90, and 135 degrees. The third row illustrates the digit recognition experiments with three patterns representing the digits 0, 1, and 8. Red and green indicated the positive and negative electrodes of the MAE, respectively.}

    \label{fig:pdf_pages}
\end{figure}

In this section, we present the experiments conducted to assess our BRC system and discuss the obtained results. Specifically, we evaluate the classification performance of the biological feature representations using three sets of experiments, each based on a different input pattern: (i) pointwise stimuli, (ii) oriented bars, and (iii) digits. Figure~\ref{fig:pdf_pages} illustrates the various input patterns on the MEA device. Table~\ref{tab:acc}, on the other hand, summarizes the classification accuracy results. To enable a direct comparison between our biological reservoir and an artificial one, we implemented an ESN with 4,096 recurrently connected rate-based units, matching the dimensionality of the feature vectors obtained from the MEA, which consists of 4,096 electrodes. The recurrent connection matrix was initialized with a sparsity of 10\%, consistent with commonly used ESN configurations and intended to reflect the sparse connectivity observed in cortical circuits. To ensure stable and rich internal dynamics, we set the spectral radius to 0.9 -- a common practice in ESN design that balances dynamic richness with fading memory. Although our reservoir state does not evolve over many time steps, and long-term memory stability was not required, this setting ensures consistent and meaningful dynamics. To mirror our experimental setup, the ESN was initialized to a resting state with all units set to zero and driven by the same external stimulus patterns used in the biological experiments. To emulate the variability of spontaneous activity observed in the biological reservoir, we introduced input noise modeled from empirical data: we analyzed $N = 25$ randomly selected \SI{10}{\milli\second} windows of spontaneous neural activity and computed the average spike count. This estimate was used to generate realistic noise, which was superimposed on the input before being fed into the ESN. After stimulus presentation, the ESN dynamics evolved for one additional time step, and the resulting state was fed to a single-layer perceptron trained to classify the input patterns. 
While the ESN serves as a useful benchmark, it should be considered an upper bound, as we expect it to perform better by design. Nevertheless, our BRC system retains the distinctive advantages of a biological substrate.

In the following, we detail the specific setups and obtained results for each set of experiments.

\begin{table}[t]
    \centering
    \small
    \caption{\textbf{Experimental Results.} Classification accuracy (light-blue column) achieved by our BRC system across the three experimental settings, corresponding to different input patterns: (i) pointwise stimuli, (ii) oriented bars, and (iii) digit recognition. For each set, we show the mean $\pm$ standard deviation across individual experiments, along with the overall average. For comparison, we also report the performance of an artificial ESN system that mirrors the same experimental setup.}
    \vspace{0.5em}
    \renewcommand{\arraystretch}{1.1}
    \setlength{\tabcolsep}{8pt}
    \newcolumntype{S}{>{\raggedright\arraybackslash}X}  
    \newcolumntype{C}{>{\centering\arraybackslash}X}    
    \begin{tabularx}{\linewidth}{S|>{\columncolor{cyan!8}}C|C}
        \toprule
        \textbf{Scenario} & \textbf{BRC Acc. (\%)} & \textbf{Artificial ESN Acc. (\%)} \\
        \midrule
        \textbf{Pointwise Stimuli} & & \\
        \arrayrulecolor{black!20}\cmidrule(lr){1-3}\arrayrulecolor{black}
        \quad Point 1 & 100\% $\pm$ 0\% & 89\% $\pm$ 13\% \\
        \quad Point 2 & 100\% $\pm$ 0\% & 71\% $\pm$ 19\% \\
        \quad Point 3 & 100\% $\pm$ 0\% & 85\% $\pm$ 18\% \\
        \quad Point 4 & 95\%  $\pm$ 9\% & 79\% $\pm$ 24\% \\
        \quad \textbf{Average} & \textbf{98\% $\pm$ 2\%} & \textbf{82\% $\pm$ 6\%} \\
        \midrule
        \textbf{Oriented Bars} & & \\
        \arrayrulecolor{black!20}\cmidrule(lr){1-3}\arrayrulecolor{black}
        \quad Bar 1 (0 degrees) & 95\% $\pm$ 9\% & 100\% $\pm$ 0\% \\
        \quad Bar 2 (45 degrees) & 91\% $\pm$ 11\% & 100\% $\pm$ 0\% \\
        \quad Bar 3 (90 degrees) & 87\% $\pm$ 27\% & 98\% $\pm$ 3\% \\
        \quad Bar 4 (135 degrees) & 95\% $\pm$ 9\% & 100\% $\pm$ 0\% \\
        \quad \textbf{Average} & \textbf{92\% $\pm$ 6\%} & \textbf{98\% $\pm$ 3\%} \\
        \midrule
        \textbf{Digit Recognition} & & \\
        \arrayrulecolor{black!20}\cmidrule(lr){1-3}\arrayrulecolor{black}
        \quad Digit 0 & 87\% $\pm$ 19\% & 100\% $\pm$ 0\% \\
        \quad Digit 1 & 100\% $\pm$ 0\% & 100\% $\pm$ 0\% \\
        \quad Digit 8 & 100\% $\pm$ 0\% & 92\% $\pm$ 10\% \\
        \quad \textbf{Average} & \textbf{95\% $\pm$ 6\%} & \textbf{97\% $\pm$ 3\%} \\
        \bottomrule
    \end{tabularx}
    \label{tab:acc}
\end{table}

\subsection{Pointwise Stimuli Experiments}
In this experimental setup, we applied pointwise input patterns by electrically stimulating single pairs of adjacent electrodes, designating one as positive and the other as negative. A simple monophasic electrical signal was employed for the stimulation in this scenario.
To select the region of interest for pointwise stimulation, we first analyzed the spontaneous network activity recorded through the MEA. Specifically, we identified four distinct electrodes in different locations that exhibited high spontaneous spike counts and selected them as positive stimulation endpoints. For each positive endpoint, the immediately adjacent electrodes were designated as negative counterparts (see also the first row of Fig.~\ref{fig:pdf_pages}). Electrical stimuli were delivered to each electrode pair with an amplitude of $A = 10\ \mu\mathrm{A}$ and a pulse width of $\delta = 20\ \mu\mathrm{s}$. 

Following the training and testing procedure, the recorded network activity was fed into a linear classifier for pattern recognition, and the classifier was trained to differentiate between the four distinct stimuli based on the spiking network activity. 
The results, shown in Tab.~\ref{tab:acc}, indicate a classification accuracy of 98\% $\pm$ 2\%, consistent with expectations given the simplicity of the task. The stimuli consisted of spatial patterns that were already linearly separable, likely contributing to the high performance. Notably, our BRC system outperforms the artificial ESN competitor, which struggles despite the simplicity of the task. We argue that this is due to the way noise is introduced in the ESN: when applied to pointwise stimuli, the noise can disrupt the signal by affecting specific input points, thereby degrading the informative content. This issue is specific to this set of experiments, as the subsequent scenarios rely on spatially distributed input patterns rather than isolated pointwise stimulation. While the simplicity of this task facilitated high accuracy, it nonetheless provided valuable confirmation that the biological reservoir responds effectively under controlled, linearly separable conditions.

\subsection{Oriented Bars Experiments}
The previous experiments with spatially distinct pointwise stimuli served as an effective preliminary stage for calibrating the stimulation protocol and validating the feasibility of linear classification as a downstream task for our BRC system. However, the simplicity of that setup, in which the patterns were linearly separable and spatially isolated, provided limited insight into the capacity of the reservoir to handle more complex and overlapping inputs.

To probe the ability of the system to discriminate among more challenging spatiotemporal patterns, we designed a second experimental scenario using stimuli composed of bar-shaped activation patterns. Specifically, we defined four stimulus categories corresponding to bars oriented at 0, 45, 90, and 135 degrees. Each pattern consists of a sequence of five pairs of adjacent electrodes (each comprising one positive and one negative endpoint) aligned along the designated orientation, with each pair spaced apart by one dilation step (see also the second row of Fig.~\ref{fig:pdf_pages}). As in the previous experiment, we employed monophasic electrical pulses.
Importantly, all four bar stimuli were presented at the same location on the MEA grid -- centered on the region exhibiting the highest spontaneous activity. This design introduces a significant challenge: different stimulus classes activate spatially overlapping regions of the neural culture, thereby eliciting partially overlapping patterns of network activity. As a result, the downstream classifier can no longer rely on simple spatial separability; instead, the BR must perform meaningful spatiotemporal transformations to map these inputs into distinct and linearly separable latent representations. The remaining aspects of the stimulation protocol and classification pipeline were kept consistent with those described previously.


The results are shown in Tab.~\ref{tab:acc}. 
The introduction of spatially overlapping bar-shaped stimuli increased the task complexity for the downstream linear classifier. The classifier achieved an average accuracy of 92\% $\pm$ 6\%, a slight decrease compared to the accuracy observed in the simpler pointwise experiment. These findings suggest that, despite the increased overlap in evoked activity across stimulus classes, the BRC system remains capable of producing distinct and linearly separable representations. This indicates the intrinsic ability of the BRC system to encode and transform input stimuli into meaningful feature embeddings, even under conditions of high spatial interference. As expected, performance was slightly lower than that of the artificial ESN, yet the BRC system maintained competitive results.


\subsection{Digit Recognition Experiments}
Finally, we evaluated the performance of the BRC system in a more complex pattern recognition task: digit recognition. In this scenario, patterns shaped like digits were defined using subsets of MEA electrodes. Specifically, we designed three distinct patterns representing the digits 0, 1, and 8, each mapped onto the MEA as \textit{electronic clock} patterns. This approach mimics the display of a digital clock, where each electrode subset corresponds to an LED segment activated to form the digit.
Specifically, for digit 1, a vertical sequence of seven electrode pairs (each consisting of a positive and a negative electrode) forms a vertical bar. The digit 0 is represented by two vertical bars of five electrode pairs spaced four electrodes apart, along with two horizontal bars at the top and bottom. Digit 8 is structured similarly to 0, but with an additional horizontal bar in the center. These input patterns are illustrated in the third row of Fig.~\ref{fig:pdf_pages}. They were selected for their high degree of overlap and because they are not linearly separable, making the classification task more challenging than in the previous scenarios.

To support this task, we adopted a more complex biphasic stimulation protocol. The stimulus amplitude was reduced to $\SI{4}{\micro\ampere}$ per electrode pair, maintaining consistent neural activation despite the lower intensity. This was feasible due to the broader distribution of the stimulation across more electrodes. Additionally, the approach improved the balance of current flow between the positive and negative endpoints, helping to preserve electrode integrity over time. As in the previous scenario, all digit patterns were presented within the same region of the MEA grid, corresponding to the area of maximal spiking activity.

Results are presented in Tab.~\ref{tab:acc}. This third experimental scenario ranks between the two previous ones, with the linear classifier achieving an accuracy of 95\% $\pm$ 6\%. Although the digit recognition task is, in principle, more challenging than the oriented bar experiments, we hypothesize that the higher performance is due to the greater number of stimulated electrodes. In the oriented bar scenario, fewer electrodes are activated, likely resulting in more localized and weaker network responses.
The artificial ESN competitor performs on average slightly better, reaching an accuracy of 97\%, as expected. Nevertheless, our BRC system delivers competitive results while retaining the inherent advantages of a biological reservoir.

\section{Conclusion and Future Works} 
\label{sec:conclusions}
In this work, we introduced an innovative approach to RC by employing a network of cultured biological neurons as the reservoir substrate. This bio-inspired paradigm presents promising prospects for enhancing energy efficiency, given the inherently lower energy consumption of biological neurons compared to artificial models. Additionally, studying biological neural networks offers valuable insights into the core mechanisms driving cognition, advancing our understanding of brain function and computation.
We successfully developed an interfacing protocol via multi-electrode array (MEA) devices and conducted an experimental investigation to evaluate the ability of the biological reservoir to recognize several input patterns. The results were promising, suggesting that biological neurons can indeed serve as a viable substrate for RC tasks, generating high-dimensional feature representations for pattern recognition. These findings lay the groundwork for further exploration of biological networks in the context of RC.

Looking ahead, we intend to expand this investigation by incorporating more complex pattern recognition tasks, allowing us to assess the scalability and robustness of the biological reservoir in a broader range of applications. Additionally, we plan to explore the integration of optogenetic stimulation protocols~\cite{meloni2020}, complementing traditional electrical stimulation, as a strategy to mitigate electrode degradation and enhance the precision and reliability of neural stimulation. This approach could lead to more sustainable and flexible neural interfacing solutions. Lastly, investigating stimulation protocols that induce synaptic plasticity will help us understand how biological learning mechanisms could be leveraged to optimize network behavior, steering it toward more task-specific and efficient responses. These future endeavors hold significant potential for advancing both the understanding and practical applications of biological reservoir computing, bringing us closer to more energy-efficient and biologically grounded cognitive computing systems.

\begin{credits}
\subsubsection{\ackname} The research was in part supported by the Matteo Caleo Foundation, by Scuola Normale Superiore (FC), by the PRIN AICult grant \#2022M95RC7 from the Italian Ministry of University and Research (MUR) (FC) and by the Tuscany Health Ecosystem - THE grant from MUR (FC, GA, ADG).

\subsubsection{\discintname}
The authors have no competing interests to declare that are relevant to the content of this article. 
\end{credits}
%
%
%
\bibliographystyle{splncs04}
\bibliography{biblio}

\end{document}